\documentclass[citeauthoryear]{llncs}
\usepackage{makeidx}
\usepackage{theapa}
\usepackage{psfig}

\renewcommand{\cite}{\citeyear}
\newcommand{\ignore}[1]{}

\begin{document}
\mainmatter

\title{Parallel Strands: A Preliminary Investigation into 
	Mining the Web for Bilingual Text}

\author{Philip Resnik}

\institute{
  Department of Linguistics and Institute for Advanced Computer Studies\\
  University of Maryland, College Park, MD 20742, USA\\
  \email{resnik@umiacs.umd.edu.edu}\\ 
  WWW home page: \texttt{http://umiacs.umd.edu/\homedir resnik/}
}

\maketitle              

\makeatletter
\renewenvironment{thebibliography}[1]
     {\section*{\refname}
      \small
      \list{}%
           {\settowidth\labelwidth{}%
            \leftmargin\parindent
            \itemindent=-\parindent
            \labelsep=\z@
            \if@openbib
              \advance\leftmargin\bibindent
              \itemindent -\bibindent
              \listparindent \itemindent
              \parsep \z@
            \fi
            \usecounter{enumiv}%
            \let\p@enumiv\@empty
            \renewcommand\theenumiv{}}%
      \if@openbib
        \renewcommand\newblock{\par}%
      \else
        \renewcommand\newblock{\hskip .11em \@plus.33em \@minus.07em}%
      \fi
      \sloppy\clubpenalty4000\widowpenalty4000%
      \sfcode`\.=\@m}
     {\def\@noitemerr
       {\@latex@warning{Empty `thebibliography' environment}}%
      \endlist}
      \def\@cite#1{#1}%
      \def\@lbibitem[#1]#2{\item[]\if@filesw
        {\def\protect##1{\string ##1\space}\immediate
      \write\@auxout{\string\bibcite{#2}{#1}}}\fi\ignorespaces}
\makeatother

\makeatletter
\def\@citeyear[#1]#2{%
    \if@filesw\immediate\write\@auxout{\string\citation{#2}}\fi%
    \edef\@citeP{}%
    \@for\@citeb:=#2\do{%
    \@ifundefined{b@\@citeb}%
     {\expandafter\def\csname b@\@citeb\endcsname{?}%
      \expandafter\def\csname Y@\@citeb\endcsname{?}%
      \@warning{Citation `\@citeb' on page \thepage\space undefined}%
     }%
     {\@ifundefined{flag@\@citeb}%
      {\global\expandafter\def\csname flag@\@citeb\endcsname{DUMMY}}%
      {}%
     }%
    \edef\B@my@dummy{\csname b@\@citeb\endcsname}%
    \if@F@cite\@F@citefalse\else{,\ }\fi%
    \csname Y@\@citeb\endcsname%
    \edef\@citeP{\csname b@\@citeb\endcsname}%
    }{\@BBN #1}%
    \@F@citetrue}
\makeatother

\begin{abstract}
Parallel corpora are a valuable resource for machine translation, but
at present their availability and utility is limited by genre- and
domain-specificity, licensing restrictions, and the basic difficulty
of locating parallel texts in all but the most dominant of the world's
languages.  A parallel corpus resource not yet explored is the World
Wide Web, which hosts an abundance of pages in parallel translation,
offering a potential solution to some of these problems and unique
opportunities of its own.  This paper presents the necessary first
step in that exploration: a method for automatically finding parallel
translated documents on the Web.  The technique is conceptually
simple, fully language independent, and scalable, and preliminary
evaluation results indicate that the method may be accurate enough to
apply without human intervention.
\end{abstract}

\section{Introduction}
\label{sec:intro}

In recent years large parallel corpora have taken on an important role
as resources in machine translation and multilingual natural language
processing, for such purposes as lexical acquisition (e.g. Gale and
Church, \cite{Gale:91b}; Melamed, \cite{melamed1997:emnlp}),
statistical translation models (e.g. Brown et al., \cite{brow1990a};
Melamed \cite{melamed1998}), and cross-language information retrieval
(e.g. Davis and Dunning, \cite{davis1995:trec4}; Landauer and Littman,
\cite{landauer1990}; also see Oard, \cite{Oard97d}).  However, for all
but relatively few language pairs, parallel corpora are available only
in relatively specialized forms such as United Nations proceedings
(LDC, \cite{ldc_homepage}), religious texts (Resnik, Olsen, and Diab,
\cite{resnik1998:chum}), and localized versions of software manuals
(Resnik and Melamed, \cite{resnik1997:anlp}).  Even for the top dozen
or so majority languages, the available parallel corpora tend to be
unbalanced, representing primarily governmental and newswire-style
texts.  In addition, like other language resources, parallel corpora
are often encumbered by fees or licensing restrictions.  For all these
reasons, following the ``more data are better data'' advice of Church
and Mercer (\cite{church/mercer1993}), abandoning balance in favor of
volume, is difficult.

A parallel corpus resource not yet explored is the World Wide Web,
which hosts an abundance of pages in parallel translation, offering a
potential solution to some of these problems and some unique
opportunities of its own.  The Web contains parallel pages in many
languages, by innumerable authors, in multiple genres and domains, and
its content is continually enriched by language change and modified by
cultural context.  In this paper I will not attempt to explore whether
such a free-wheeling source of linguistic content is better or worse
than the more controlled parallel corpora in use today.

Rather, this paper presents the necessary first step in that
exploration: a method for automatically finding parallel translated
documents on the Web that I call STRAND ({\bf S}tructural {\bf
T}ranslation {\bf R}ecognition for {\bf A}cquiring {\bf N}atural {\bf
D}ata).  The technique is conceptually simple, fully language
independent, and scalable, and preliminary evaluation results indicate
that the method may be accurate enough to apply without human
intervention.

In Section~\ref{sec:architecture} I lay out the STRAND architecture
and describe in detail the core of the method, a language-independent
structurally based algorithm for assessing whether or not two Web
pages were intended to be parallel translations.
Section~\ref{sec:testing} presents preliminary evaluation, and
Section~\ref{sec:future} discusses future work.

\section{The STRAND Architecture}
\label{sec:architecture}

\begin{figure}[t]
\hbox{
\centerline{
\psfig{figure=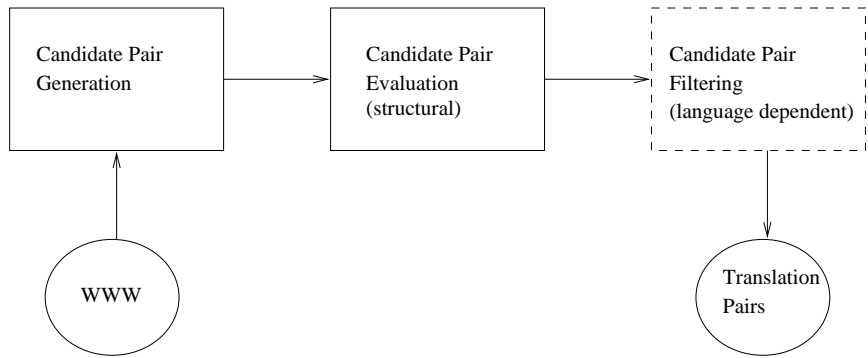,width=4.5in}}
}
\caption{The STRAND architecture \label{fig:strand}}
\end{figure}
As Figure~\ref{fig:strand} illustrates, the STRAND architecture is a simple
pipeline.  Given a particular pair of languages of interest, a {\em
candidate generation} module first generates pairs
$\langle$url1,url2$\rangle$ identifying World Wide Web pages that may be
parallel translations.\footnote{A URL, or {\em uniform resource locator},
is the address of a document or other resource on the World Wide Web.}
Next, a language independent {\em candidate evaluation} module behaves as a
filter, keeping only those candidate pairs that are likely to actually be
translations.  Optionally, a third module for {\em language-dependent
filtering} applies additional filtering criteria that might depend upon
language-specific resources.  The end result is a set of candidate pairs
that can reliably be added to the Web-based parallel corpus for these two
languages.

The approach to candidate evaluation taken in this paper has a useful side
effect: in assessing the likelihood that two HTML documents are parallel
translations, the module produces a segment-level alignment for the
document pair, where segments are chunks of text appearing in between
markup.  Thus STRAND has the potential of producing a segment-aligned
parallel corpus rather than, or in addition to, a document-aligned parallel
corpus.  In this paper, however, only the quality of document-level
alignment is evaluated.\footnote{HTML, or {\em hypertext markup language},
is currently the authoring language for most Web pages.  The STRAND
approach should also be applicable to SGML, XML, and other formats, but
they will not be discussed here.}

\subsection{Candidate Generation} 
\label{sec:generation}

\begin{figure}[t]
\hbox{
\centerline{
\framebox{\psfig{figure=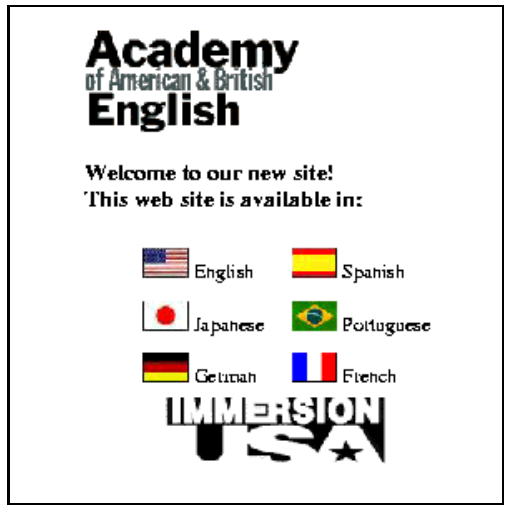,width=2.5in}}}
}
\caption{A page containing links to parallel translations 
	\label{fig:academy}}
\end{figure}

At present the candidate generation module is implemented very simply.
First, a query is submitted to the Altavista Web search engine, which
identifies Web pages containing at least one hyperlink where
'language1' appears in the text or URL associated with the link, and
at least one such link for language2.\footnote{An ``anchor'' is a
piece of HTML document that encodes a hypertext link.  It typically
includes the URL of the page being linked to and text the user can
click on to go there; it may contain other information, as well.  The
URL for Altavista's ``advanced search'' page is
http://altavista.digital.com/cgi-bin/query?pg=aq\&what=web.}  For
example, Altavista's ``advanced search'' can be given Boolean queries
in this form:
\begin{verbatim}
  anchor:"language1"  AND  anchor:"language2"
\end{verbatim}
A query of this kind, using {\em english\/} and {\em french\/} as
language1 and language2, respectively, locates the home page of the Academy
of American \& British English, at http://www.academyofenglish.com/
(Figure~\ref{fig:academy}), among many others.
\ignore{
For example, the 'hits' included in the Altavista search results will
include pages containing oft-seen sets of links like the following:

\begin{verbatim}
  English | French | German | Spanish
\end{verbatim}
} 

On some pages, images alone are used to identify alternative language
versions --- the flag of France linking to a French-language page, for
example, but without the word ``French'' being visible to the user.
Text-based queries can still locate such pages much of the time,
however, because the HTML markup for the page conventionally includes
the name of the language for display by non-graphical browsers (in the
ALT field of the IMG element).  Names of languages sometimes also
appear in other parts of a URL --- for example, the file containing
the image of the French flag might be named {\tt french.gif}.  The
Altavista query above succeeds in identifying all these cases and
numerous others.

In the second step of candidate generation, each page returned by Altavista
is automatically processed to extract all pairs $\langle$url1,url2$\rangle$
appearing in anchors $\langle a_1,a_2 \rangle$ such that $a_1$ contains
`language1', $a_2$ contains `language2', and $a_1$ and $a_2$ are no more
than 10 lines apart in the HTML source for the page.  This distance
criterion captures the fact that for most Web pages that point off to
parallel translations, the links to those translations appear relatively
close together, as is the case in Figure~\ref{fig:academy}.

I have not experimented much with variants on this simple method for
candidate generation, and it clearly could be improved in numerous
ways to retrieve a greater number of good candidates. For example, it
might make sense to issue a query seeking documents in language2 with
an anchor containing `language1' (e.g. query Altavista for
pages in French containing pointers to `English', to
capture the many pairs connected by a link saying `English
version').  Or, it might be possible to exploit parallel URL
and/or directory structure; for example, the URLs
http://amta98.org/en/program.html and
http://amta98.org/fr/program.html are more likely than other URL pairs
to be referring to parallel pages, and the directory subtrees under
{\em en} and {\em fr} on the fictitious amta98.org server might be
well worth exploring for other potential candidate pairs.

For this initial investigation, however, generating a reasonable set
of candidates was the necessary first step, and the simple approach
above works well enough.  Alternatives to the current candidate
generation module will be explored in future work.

\subsection{Candidate Evaluation}
\label{sec:evaluation}

The core of the STRAND approach is its method for evaluating candidate
pairs --- that is, determining whether two pages should be considered
parallel translations.  This method exploits two facts.  First,
parallel pages are filled with a great deal of identical HTML markup.
Second, work on bilingual text alignment has established that there is
a reliably linear relationship in the lengths of text translations
(Gale and Church, \cite{Gale:91a}; Melamed, \cite{mel96a}).  The
algorithm works by using pieces of identical markup as reliable points
of correspondence and computing a best alignment of markup and
non-markup chunks between the two documents.  It then computes the
correlation for the lengths of the non-markup chunks.  A test for the
significance of this correlation is used to decide whether or not a
candidate pair should be identified as parallel text.  

\begin{figure}[t]
\hbox{
\centerline{
\framebox{\psfig{figure=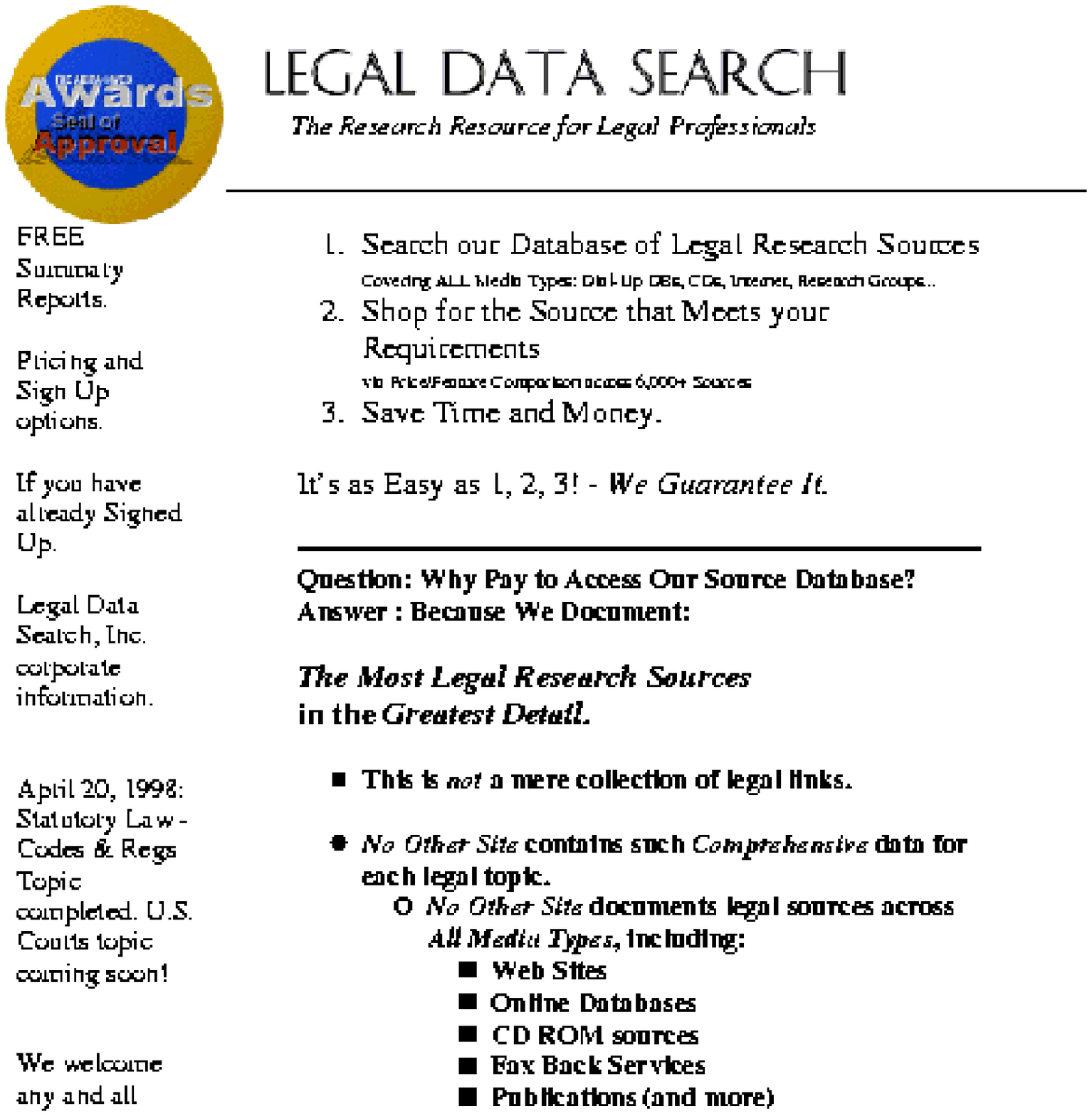,width=2.35in,height=2.8in}}
\framebox{\psfig{figure=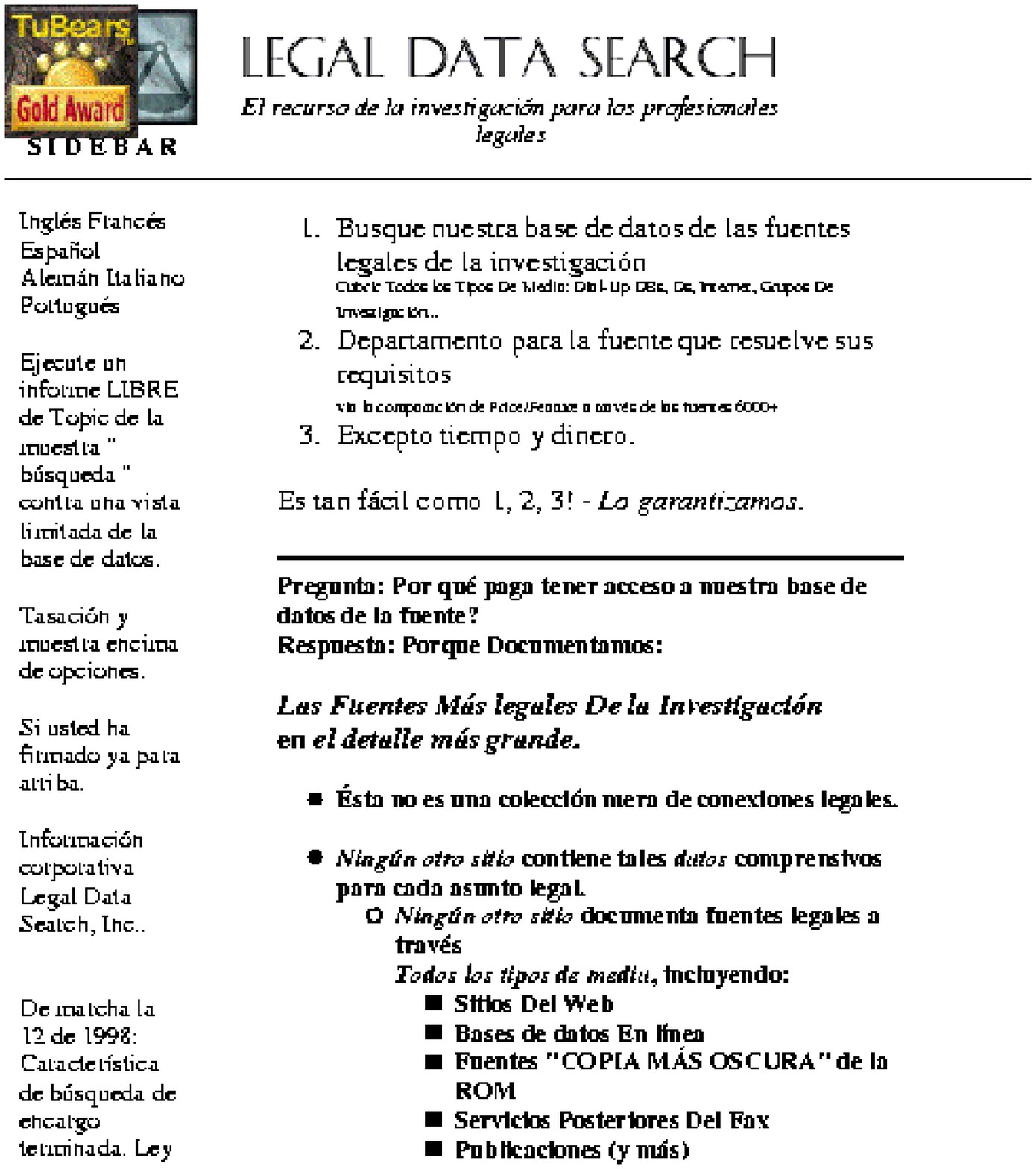,width=2.35in,height=2.8in}}}
}
\caption{Example of a candidate pair \label{fig:eval_pages}}
\end{figure}

For example, Figure~\ref{fig:eval_pages} shows fragments from a pair of
pages identified by STRAND's candidate generation module in the experiment
to be described in Section~\ref{sec:testing}.
An English page is at left, Spanish at right.\footnote{Source:
http://www.legaldatasearch.com/.} Notice the extent to which the page
layout is parallel, and the way in which corresponding units of text
--- list items, for example --- have correspondingly greater or
smaller lengths.

In more detail, the steps in candidate evaluation are as follows:

\begin{enumerate}
\item Linearize.  Both documents in the candidate pair are run through
a markup analyzer that acts as a transducer, producing a linear
sequence containing three kinds of token:

\begin{center}
\begin{tabular}{lcl}
   \verb|[START:element_label]|  & & e.g. \verb|[START:A]|, \verb|[START:LI]| \\
   \verb|[END:element_label]|    & & e.g. \verb|[END:A]|  \\
   \verb|[Chunk:length]|         & & e.g. \verb|[Chunk:174]| \\ 
\end{tabular}
\end{center} 

\item Align the linearized sequences.  There are many approaches one
can take to aligning sequences of elements.  In the current prototype,
the Unix {\em sdiff} utility does a fine job of alignment, matching up
identical START and END tokens in the sequence and Chunk tokens of
identical length in such a way as to minimize the differences between
the two sequences.  For example, consider two documents that begin as
follows:

\begin{center}
\begin{tabular}{l|cl}
$<$HTML$>$ 			      & & $<$HTML$>$ \\
$<$TITLE$>$Emergency Exit$<$/TITLE$>$ & & $<$TITLE$>$Sortie de Secours$<$/TITLE$>$ \\
$<$BODY$>$			      & & $<$BODY$>$ \\
$<$H1$>$Emergency Exit$<$/H1$>$       & & Si vous \^etes assis \`a \\
If seated at an exit and              & & c\^ot\'e d'une... \\
$\vdots$                              & & $\vdots$
\end{tabular}
\end{center}

The aligned linearized sequence would be as follows:\footnote{Note that
whitespace is ignored in counting chunk lengths.}

\begin{verbatim}
          [START:HTML]          [START:HTML] 
          [START:TITLE]         [START:TITLE] 
          [Chunk:12]            [Chunk:15] 
          [END:TITLE]           [END:TITLE] 
          [START:BODY]          [START:BODY] 
          [START:H1]           
          [Chunk:12]           
          [END:H1]           
          [Chunk:112]           [Chunk:122]
\end{verbatim} 

\item Threshold the aligned, linearized sequences based on mismatches.
When two pages are not parallel, there is a high proportion of
mismatches in the alignment --- sequence tokens on one side that have
no corresponding token on the other side, such as the tokens
associated with the H1 element in the above example.  This can happen,
for example, when two documents are translations up to a point,
e.g. an introduction, but one document goes on to include a great deal
more content than another.  Even more frequently, the proportion is
high when two documents are {\em prima facie} bad candidates for a
translation pair.  For these reasons, candidate pairs whose mismatch
proportion exceeds a constant, K, are eliminated at this stage.  My
current value for K was set manually at 20\% based on experience with
a development set, and that value was frozen and used in the
experiment described in the next section.  In that experiment
evaluation of STRAND was done using a different set of previously
unseen documents, for a different language pair than the one used
during development. \\

\item Compute a confidence value. Let $\langle X,Y \rangle =
\{(x_1,y_1), \ldots, (x_n,y_n)\}$ be the lengths for the aligned Chunk
tokens in Step 2, such that $x_j$ is not equal to $y_j$. (When they
are exactly equal, this virtually always means the aligned segments
are not natural language text.  If included these inflate the
correlation coefficient.)  For the above alignment this would be
$\{(12,15), (112,122), \ldots\}$.  Compute the Pearson correlation
coefficient r(X,Y), and compute the significance of that correlation
in textbook fashion.  Note that the significance calculation takes the
number {\em n} of aligned text segments into account.  The resulting
{\em p} value is used to threshold significance: using the standard
threshold of $p < .05$ (i.e. 95\% confidence that the
correlation would not have been obtained by chance) worked well during
development, and I retained that threshold in the evaluation described
in the section that follows.
\end{enumerate}

\begin{figure}[t]
\hbox{
\centerline{
\psfig{figure=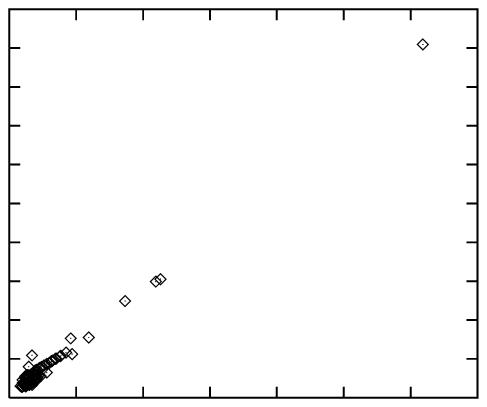,width=1.6in,height=1.6in}
\psfig{figure=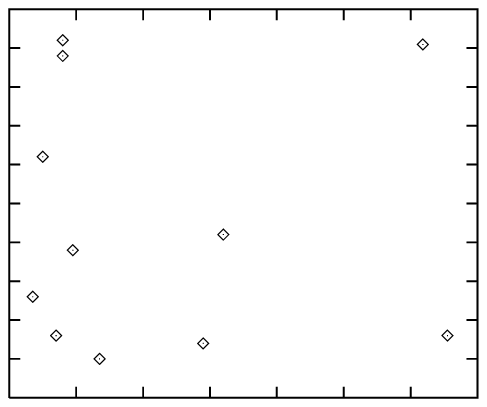,width=1.6in,height=1.6in}
\psfig{figure=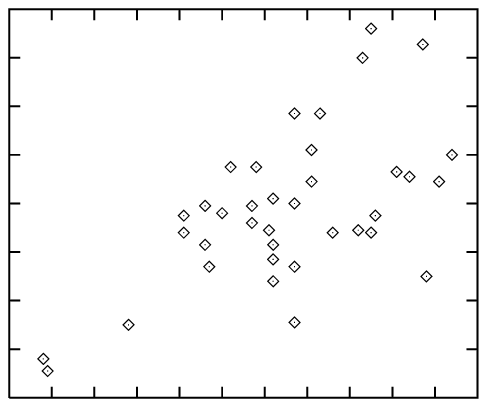,width=1.6in,height=1.6in}}
}
\caption{Scatterplots illustrating reliable correlation in lengths of aligned
segments for good translation pairs (left and right), and lack of
correlation for a bad pair (center). \label{fig:correl}}
\end{figure}

\noindent Figure~\ref{fig:correl} shows plots of $\langle X,Y\rangle$
for three real candidate pairs.  \ignore{18 (good p, low r: proverbs),
98 (good: legaldatasearch), 158 (bad: IDP)}
At left is the pair illustrated in Figure~\ref{fig:eval_pages}, correctly
accepted by the candidate evaluation module with $\mbox{r} = .99, \mbox{p}
< .001$.  At center is a pair correctly rejected by candidate evaluation;
in this case $\mbox{r} = .24, \mbox{p} > .4$, and the mismatch proportion
exceeds 75\%.  And at right is another pair correctly accepted; in this
more unusual case, the correlation is lower ($\mbox{r} = .57$) but
statistically very reliable because of the large number of data 
points ($\mbox{p} < .0005$).

Notice that a by-product of this structurally-driven
candidate evaluation scheme is a set of aligned Chunk tokens.  These
correspond to aligned non-markup segments in the document pair.
Evaluating the accuracy of this segment-level alignment is left for
future work.

\subsection{Language-Dependent Filtering}
\label{sec:dependent}

I have not experimented with further filtering of candidate pairs
since, as shown in the next section, precision is already quite
high. However, experience with the small number of false positives I
have seen suggests that automatic language identification on the
remaining candidate pairs might weed out the few that remain.  Very
high accuracy language identification using character {\em n}-gram
models requires only a modest amount of training text known to be in
the languages of interest (Dunning, \cite{dunning1994:language_id};
Grefenstette, \cite{grefenstette1995}).

\section{Evaluation}
\label{sec:testing}

I developed the STRAND prototype using English and French as the
relevant pair of languages.  For evaluation I froze the code and all
parameters and ran the prototype for English and Spanish, not having
previously looked at English/Spanish pairings on the Web.

For the candidate generation phase, I followed the approach of
Section~\ref{sec:generation} and generated candidate document pairs from
the first 200 hits returned by the Altavista search engine, leading to a
set of 198 candidate pairs of URLs that met the distance criterion.

Of those 198 candidate pairs, 12 were pairs where url1 and url2
pointed to identical pages, and so these are eliminated from
consideration.  In 96 cases one or both pages in the pair could not be
retrieved (page not found, moved, empty, server unreachable, etc.).
The remaining 90 cases are considered the set of candidate pairs for
evaluation.

I evaluated the 90 candidate pairs by hand, determining that 24
represented true translation pairs.\footnote{A few of the 90 candidate
pairs were encoded in non-HTML format, e.g. PDF ({\em portable
document format}).  I excluded these from consideration {\em a
priori\/} because STRAND's capabilities are currently limited to
HTML.}  The criterion for this determination was the question: Was
this pair of pages intended to provide the same content in the two
different languages?  Although admittedly subjective, the judgments
are generally quite clear; I include URLs in an on-line Appendix so
that the reader may judge for himself or herself.  The STRAND
prototype's performance against this test set was as follows:

\begin{itemize}
\item The candidate evaluation module identified 17 of the 90
candidate pairs as true translations, and was correct for 15 of those
17, a precision of 88.2\%.  (A language-dependent filtering module
with 100\% correct language identification would have eliminated one
of the two false positives, giving a precision of 93.8\%.  However,
language-dependent filtering was not used in this evaluation.) \\
\item The algorithm identified 15 of 24 true translation
pairs, a recall of~62.5\%.
\end{itemize}

\noindent Manual assessment of the translation pairs retrieved by the
algorithm suggests that they are representative of what one would expect to
find on the Web: the pages vary widely in length, content, and the
proportion of usable parallel natural language text in comparison to
markup, graphics, and the like.  However, I found the yield of genuine
parallel text --- content in one language and its corresponding translation
in the other --- to be encouraging.  The reader may form his or her own
judgment by looking at the pages identified in the on-line Appendix.

\section{Future Work}
\label{sec:future}

At present it is difficult to estimate how many pairs of translated
pages may exist on the World Wide Web.  However, it seems fair to say
that there are a great many, and that the number will increase as the
Web continues to expand internationally.  The method for candidate
generation proposed in this paper makes it possible to quickly locate
candidate pairs without building a Web crawler, but in principle one
could in fact think of the entire set of pages on the Web as a source
for candidate generation.  The preliminary figures for recall and
especially for precision suggest that large parallel corpora can be
acquired from the Web with only a relatively small degree of noise,
even without human filtering.  Accurate language-dependent filtering
(e.g. based on language identification, as in
Section~\ref{sec:dependent}) would likely increase the precision,
reducing noise, without substantially reducing the recall of useful,
true document pairs.  In addition to language-dependent filtering, the
following are some areas of investigation for future work.

\begin{itemize}
\item {\bf Additional evaluation.}  As advertised in the title of this
paper, the results thus far are preliminary.  The STRAND approach
needs to be evaluated with other language pairs, on larger candidate
sets, with independent evaluators being used in order to accurately
estimate an upper bound on the reliability of judgments as to whether
a candidate pair represents a true translation.  One could also
evaluate how precision varies with recall, but I believe for this task
there are sufficiently many genuine translation pairs on the Web and a
sufficiently high recall that the focus should be on maximizing
precision.  Alternative approaches to candidate generation from the
Web, as discussed in Section~\ref{sec:generation}, are a topic for
further investigation. \\

\item {\bf Scalability.}  The prototype, implemented in decidedly
non-optimized fashion using a combination of perl, C, and shell
scripts, currently evaluates candidate pairs at approximately 1.8
seconds per candidate on a Sun Ultra 1 workstation with 128 megabytes
of real memory, when the pages are already resident on a disk on the
local network (though not local to the workstation itself).  Thus,
excluding retrieval time of pages from the Web, evaluating 1~million
retrievable candidate pairs using the existing prototype would take
just over 3~weeks of real time.  However, STRAND can easily be run in
parallel on an arbitrary number of machines, and the prototype
reimplemented in order to obtain significant speed-ups.  The main
bottleneck to the approach, the time spent retrieving pages from the
Web, is still trivial if compared to manual construction of corpora.
In real use, STRAND would probably be run as a continuous process,
constantly extending the corpus, so that the cost of retrieval would
be amortized over a long period. \\

\item {\bf Segment alignment.} As discussed in
Section~\ref{sec:evaluation}, a by-product of the candidate evaluation
module in STRAND is a set of aligned text segments.  The quality of
the segment-level alignment needs to be evaluated, and should be
compared against alternative alignment algorithms based on the
document-aligned collection. \\

\item {\bf Additional filtering.}  Although a primary goal of this
work is to obtain a large, heterogeneous corpus, for some purposes it
may be useful to further filter document pairs.  For example, in some
applications it might be important to restrict attention to document
pairs that conform to a particular genre or belong to a particular
topic.  The STRAND architecture of Figure~\ref{fig:strand} is clearly
amenable to additional filtering modules such as document
classification incorporated into, or pipelined with, the
language-dependent filtering stage. \\

\item {\bf Dissemination.}  Although text out on the Web is generally
intended for public access, it is nonetheless protected by copyright.
Therefore a corpus collected using STRAND could not legally be
distributed in any straightforward way.  However, legal constraints do
not prevent multiple sites from running their own versions of STRAND,
nor any such site from distributing a list of URLs for others to
retrieve themselves.  Anyone implementing this or a related approach
should be careful to observe protocols governing automatic programs
and agents on the Web.\footnote{See
http://info.webcrawler.com/mak/projects/robots/robots.html.} \\
\end{itemize}

The final and most interesting question for future work is: What can
one {\em do} with a parallel corpus drawn from the World Wide Web?  I
find two possibilities particularly promising.  First, from a
linguistic perspective, such a corpus offers opportunities for
comparative work in lexical semantics, potentially providing a rich
database for the cross-linguistic realization of underlying semantic
content.  From the perspective of applications, the corpus is an
obvious resource for acquisition of translation lexicons and
distributionally derived representations of word meaning.  Most
interesting of all, each possibility is linked to many others,
seemingly without end --- much like the Web itself.

\section*{Acknowledgments}

This work was supported in part by DARPA/ITO contract
N66001-97-C-8540, Department of Defense contract MDA90496C1250, and a
research grant from Sun Microsystems Laboratories.  I am grateful to
Dan Melamed, Doug Oard, and David Traum for useful discussions.

\section*{Appendix: Experimental Data}

At URL http://umiacs.umd.edu/\homedir resnik/amta98/amta98\_appendix.html
the interested reader can find an on-line Appendix containing the
complete test set described in Section~\ref{sec:testing}, with
STRAND's classifications and the author's judgments.

\end{document}